\documentclass[review]{elsarticle}

\usepackage{lineno,hyperref}
\modulolinenumbers[5]
\usepackage{threeparttable}
\usepackage{arydshln}
\usepackage{setspace}
\usepackage{verbatimbox}

\begin{document}

\begin{frontmatter}

\title{Prior-enlightened and Motion-robust Video Deblurring}


\author[mymainaddress]{Ya Zhou}

\author[mysecondaryaddress]{Jianfeng Xu}

\author[mysecondaryaddress]{Kazuyuki Tasaka}

\author[mymainaddress]{Zhibo Chen\corref{mycorrespondingauthor}}
\cortext[mycorrespondingauthor]{Corresponding author}
\ead{chenzhibo@ustc.edu.cn}

\author[mymainaddress]{Weiping Li}

\address[mymainaddress]{CAS Key Laboratory of Technology
	in Geo-Spatial Information Processing and Application System,
	University of Science and Technology of China, Hefei 230026, China}
\address[mysecondaryaddress]{KDDI research, Inc., Fujimino-shi, Saitama, 356-8502 Japan}

\begin{abstract}
Various blur distortions in video will cause negative impact on both human viewing and video-based applications, which makes motion-robust deblurring methods urgently needed. Most existing works have strong dataset dependency and limited generalization ability in handling challenging scenarios, like blur in low contrast or severe motion areas, and non-uniform blur. Therefore, we propose a PRiOr-enlightened and MOTION-robust video deblurring model (PROMOTION) suitable for challenging blurs. On the one hand, we use 3D group convolution to efficiently encode heterogeneous prior information, explicitly enhancing the scenes' perception while mitigating the output's artifacts. On the other hand, we design the priors representing blur distribution, to better handle non-uniform blur in spatio-temporal domain. Besides the classical camera shake caused global blurry, we also prove the generalization for the downstream task suffering from local blur. Extensive experiments demonstrate we can achieve the state-of-the-art performance on well-known REDS and GoPro datasets, and bring machine task gain.
\end{abstract}

\begin{keyword}
Video deblurring\sep prior\sep motion-robust\sep hand pose estimation
\end{keyword}

\end{frontmatter}


\section{Introduction}

The popularity of mobile devices makes it easier for users to create videos. However, in general, the inevitable device shake during shooting usually produces an annoying visual experience. As a result, global motion blur is increasingly considered by device manufacturers in the development of new products. As one of the solutions to this, a robust video deblurring method can often be designed to adapt to different shooting scenarios.

On the other hand, thanks to deep learning, more and more video-based downstream tasks, such as hand pose estimation~\cite{handpose1,learningjoint,handpose2}, face recognition~\cite{facerecog1,facerecog2}, and ReID~\cite{reid3,reid4}, have experienced significant improvements in the last few years. However, many of the existing deep learning-based approaches are end-to-end, which often require high-quality input data for more accurate estimation or identification. In these tasks, cameras are often fixed, like monitoring scenario, and only some local areas are moving in the picture. Naturally, this requires a deblurring method that is sensitive to local blur to preprocess the input source in preparation for subsequent tasks. 

Inspired by the needs in various deblurring scenarios, this paper would like to propose a motion-robust video deblurring method that is effective for both global and local blur. In recent years, more and more deblurring works have emerged. In terms of image deblurring, a major trend is to use multi-scale structure~\cite{Nah_2017_CVPR,Tao_2018_CVPR,multiscaleprior}, or use pyramid structure~\cite{deblurganv2}, for the purpose of enabling the network to have varying receptive fields, so as to effectively deal with different degrees of blur. For more complex video deblurring, in addition to conventionally aligning multiple frames and deblurring the center frame like EDVR~\cite{edvr} and DeBlurNet~\cite{deblurnet}, it is often focused on how to make full use of inter-frame redundancy, such as using recurrent structure~\cite{recurrentKim_2017_ICCV,recurrentZhang_2018_CVPR,recurrentNah_2019_CVPR}, 3D convolution~\cite{adversarialSTlearning,anempirical}, or encoding each frame separately and then aggregating them to decode~\cite{ntire}. Therefore, research on the utilization of multi-scale receptive fields and inter-frame information can be considered relatively mature. Different from these research perspectives, we study the importance of prior information of data itself to video deblurring. 

Deblurring challenging blurs, including blur in low-contrast and severe motion (e.g., close to cameras) areas, non-uniform blur (e.g., local blur), and other special cases, has been not well solved by existing video deblurring yet is a very important problem. This inspires us to dig out the inherent information of scenes themselves when designing priors, and try to make our design plug-and-play and easy to migrate. 

As the typical representative of conventional methods, EDVR mainly consists of predeblur, alignment, fusion, and reconstruction modules, as the golden background area shown in Fig.~\ref{fig:framework}. Among them, predeblur module is essentially a pyramid network composed of residual blocks, rather than any existing deblurring work. It is inserted before alignment module to improve alignment accuracy. Then, in the alignment module, each neighboring frame is aligned to the center one at feature level. The temporal and spatial attention (TSA) fusion module next assigns pixel-level aggregation weights on each frame according to correlation, and further fuses image information of different frames. The fused feature then passes through a reconstruction module to restore the center frame.

\begin{figure}[!t]
	\begin{center}
		\includegraphics[width=1.0\columnwidth]{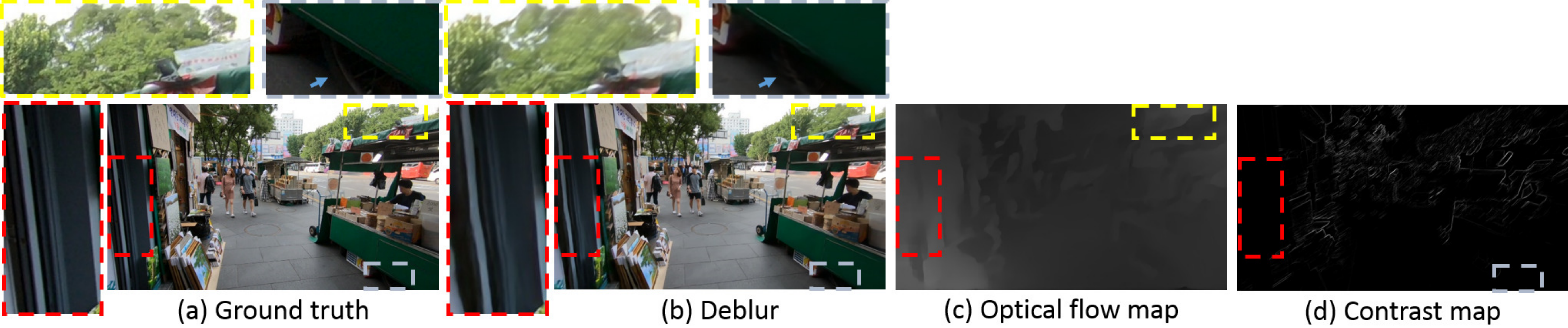} 
	\end{center}
	\caption{Illustration of the weaknesses of the typical conventional method EDVR~\cite{edvr}, including structure distortion, weak deblurring performance on large optical flow or low contrast areas. Since it lacks the effective constraint of heterogeneous prior information.}
	\label{fig:observation1}
\end{figure}

\begin{figure}[!t]
	\begin{center}
		\includegraphics[width=1.0\columnwidth]{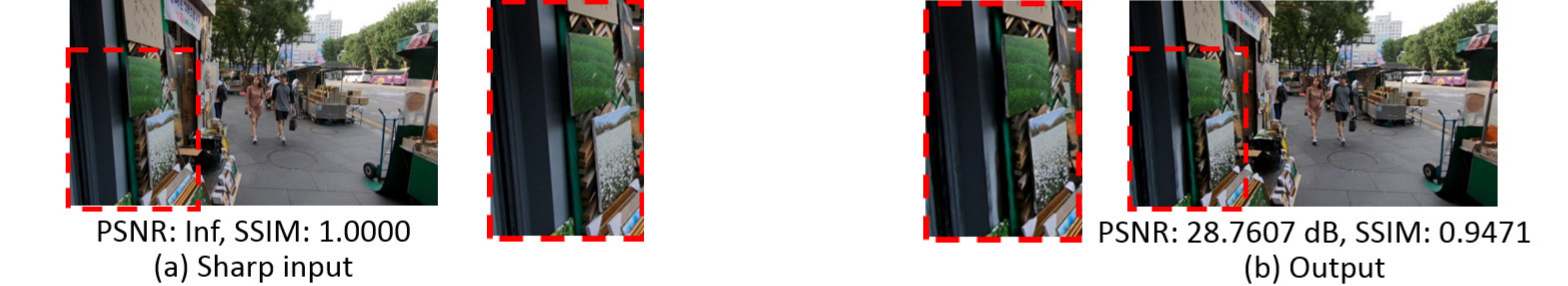} 
	\end{center}
	\caption{Illustration that EDVR seriouly worsens the sharp input since it cannot discriminate the sharp information.} 
	\label{fig:observation2}
\end{figure}

However, by analyzing its design and deblurring results, we have some interesting observations: 
\begin{enumerate}[(1)]
\item \textbf{Fail to effectively and deeply utilize temporal correlation to help deblur.} The aligned feature is obtained by just concatenating. This lacks the mining of motion information. On the other hand, 2D convolution also has limited capability in modelling long-term temporal dependency. 
\item \textbf{The deblurred frames are prone to be infected with structure distortion}, like straight lines becoming twisty, as the red dashed boxes shown in Fig.~\ref{fig:observation1} (a) and (b). This is because the model does not have the constraint of structure preserving, whatever in model and loss design.
\item \textbf{The larger the optical flow, the more difficult it is to eliminate blur.} We observe the red and yellow blocks presented in Fig.~\ref{fig:observation1} (a)(b)(c). It is not difficult to find these areas with large motion have a poor recovery of detail, such as the leaves. This is because the model lacks the prior information of motion, which makes it unable to adjust the intensity of deblurring adaptively. 
\item \textbf{The lower the contrast, the harder it is to remove blur.} Similarly as the red and blue boxes shown in Fig.~\ref{fig:observation1} (a)(b)(d), the low contrast areas on the door edge and the tire are not well restored, since the model lacks the ability to perceive the contrast distribution of scenes.
\item \textbf{The original sharp information cannot be preserved.} For example in Fig.~\ref{fig:observation2}, if we input the sharp frame mixed with blur frames, the output is seriously worse than the input, which indicates the model has no ability to distinguish sharp and blurred inputs (non-uniform blur). 
\item \textbf{The loss design}, that is, Charbonnier loss~\cite{charbonnierloss} which calculates the error of each pixel indiscriminately, \textbf{cannot also effectively restrain artifacts and embody the influence from non-uniform blur.} 
\end{enumerate}

Based on the above analysis, we enhance the model's perception about scenes from the use of prior information, the way of feature extraction, and the design of constraints. Our contributions are as follows:
\begin{itemize}
\item We utilize 3D group convolution to encode heterogeneous prior information to explicitly supplement the model's multidimensional perception about scenes, which is not learned well before, due to out of consideration both in model and loss design. The heterogeneous priors involve structure, motion, and contrast, which effectively constrain the artifacts of the model and enhance the detail recovery of challenging blurs. This corresponds to observation (1) to (4).

\item Both in the temporal and spatial dimension, we increase the model's ability to distinguish sharp and blurred inputs. In temporal dimension, the blur reasoning vector is embedded in aligned feature to represent the blur degree of each frame. In spatial dimension, optical flow based attention is used to indicate the spatial blur degree within single frame. This corresponds to observation (5).

\item The dual loss function, considering from pixel level and perceptual level simultaneously, is designed to effectively constrain the blur removal and guarantee subjective quality. This corresponds to observation (6).

\begin{figure}[!t]
	\begin{center}
		\includegraphics[width=1.0\linewidth]{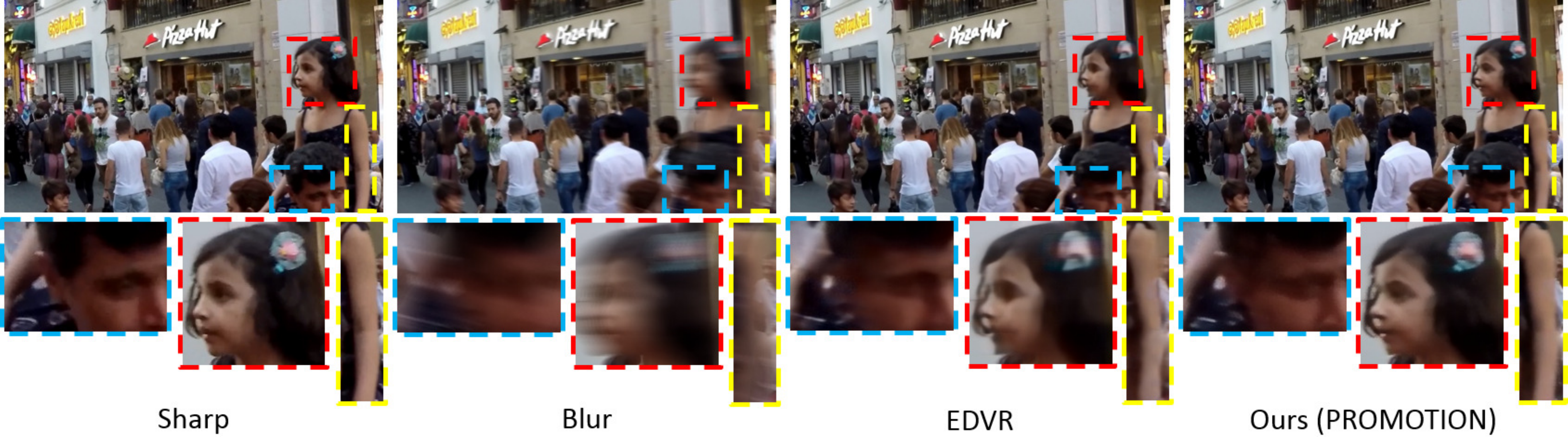} 
	\end{center}
	\caption{Performance illustration of our method. Compared with EDVR~\cite{edvr} which has limited generalization ability of handling challenging blurs and tends to introduce structural distortion, due to the utilization of priors, our method can deliever more visual pleasing results, especially with more detail recovery and better fidelity in structure.}
	\label{fig:comp1}
\end{figure}

\item  A video deblurring dataset for hand pose estimation is firstly constructed. Then we verified that the estimation accuracy is improved after deblurring. This indirectly reflects the effectiveness for local blurry scenes, and further indicates our model can bring not only visual gain but also machine taks gain.
\end{itemize}

The use of priori information brings benefits to video deblurring task, especially in terms of subjective effect, as shown in Fig.~\ref{fig:comp1} where our results have more detail restoration and structure fidelity.

\section{Related work}

\noindent\textbf{Video deblurring.} 
As one of the first treatments of blur in conjunction with optical flow,~\cite{tra_deblur1} can calculate accurate flow in the presence of spatially-varying motion blur. Instead of assuming the captured scenes are static,~\cite{tra_deblur2} first proposes an effective general video deblurring method for dynamic scenes.~\cite{tra_deblur3} exploits semantic segmentation to undertand the scene contents and further presents a pixel-wise non-linear kernel model to explain motion blur. 
Since the first end-to-end data-driven video deblurring method DeBlurNet was propopsed~\cite{deblurnet}, in the past two years, the success of deep learning has brought significant promotion to video deblurring~\cite{adversarialSTlearning,anempirical,recurrentKim_2017_ICCV,recurrentZhang_2018_CVPR,recurrentNah_2019_CVPR,reblur2deblur,stfan,rdurud,deblurnet,anempirical,edvr}. Different from image deblurring only focusing on the mining of spatial information, such as using multi-scale receptive field~\cite{Nah_2017_CVPR,Tao_2018_CVPR,multiscaleprior}, video deblurring also needs to focus on how to effectively use redundant information in the temporal domain to assist deblurring. 

The conventional approach is to stack multiple frames together as a single input and then uses 2D convolutions to extract features~\cite{deblurnet,anempirical,edvr}. Compared to the classic DeBlurNet~\cite{deblurnet}, \cite{anempirical} has more consideration about the complexity and parameters of the model. Before going through 2D convolutions, EDVR aligns adjacent frames to better aggregate information~\cite{edvr}.

Some works use recurrent neural networks (RNNs) for sequential data processing~\cite{recurrentKim_2017_ICCV,recurrentZhang_2018_CVPR,recurrentNah_2019_CVPR}. \cite{recurrentKim_2017_ICCV} designs a spatio-temporal recurrent network which extends the receptive field while keeping the network small, and uses dynamic temporal blending to enforce temporal consistency. \cite{recurrentNah_2019_CVPR} presents a RNN-based video deblurring method that exploits both the intra-frame and inter-frame recurrent schemes and updates the hidden state multiple times internally during a single time-step. Another spatially variant RNN for dynamic scene deblurring is proposed in~\cite{recurrentZhang_2018_CVPR}, where the weights of the RNN are learned by a deep CNN. 

Another part studies how to extract pixel-wise information to handle the spatially variant blur~\cite{reblur2deblur,stfan,rdurud}. Similar to CycleGAN's circular idea~\cite{cyclegan}, \cite{reblur2deblur} first deblurs each frame separately, then estimates optical flow and pixel-wise blur kernels to reblur the estimated sharp images, which makes the network fine-tuned via self-supervised learning. Similarly, \cite{stfan} proposes a spatial-temporal network for video deblurring based on filter adaptive convolutional layers, and the network is able to dynamically generate element-wise alignment and deblurring filters in order. \cite{rdurud} presents a motion deblurring kernel learning network that predicts the per-pixel deblur kernel and a residual image with two novel base blocks named residual down-up and residual up-down blocks. 

In addition, 3D convolution has been also used for video deblurring recently to extract spatio-temporal information simultaneously~\cite{adversarialSTlearning}. \cite{adversarialSTlearning} applies 3D convolutions to RGB frames, and uses a discriminator for adversarial training. While we use 3D convolution for priori frames, and furthermore, we realize the fusion of heterogeneous priors by means of group convolution. 

\noindent\textbf{Priori information.}
In human action recognition, \cite{3dconv} designs a set of hardwired kernels in the network to encode the prior knowledge, including gray, gradient and optical flow in horizontal and vertical directions, which is proved to usually lead to better performance as compared to the random initialization. In image deblurring, there are also some works using different prior information~\cite{darkchannelprior,extremechannelprior,deblurtext,multiscaleprior,discriminativeprior,classprior,featureprior}. Inspired by the observation that the dark channel of blurred images is less sparse, \cite{darkchannelprior} presents a blind image deblurring method based on the dark channel prior. Similarly, \cite{extremechannelprior} defines the bright channel prior and takes advantage of both bright and dark channel prior to deblur images. In~\cite{deblurtext}, a $L_{0}$-regularized prior based on intensity and gradient is designed for text image deblurring. \cite{multiscaleprior} proposes a blind image deblurring method using multi-scale latent structure prior. Recently, \cite{discriminativeprior} formulates the image prior as a binary classifier, that is whether the image is clear or not, and presents the deblurring method based on a data-driven discriminative prior. And~\cite{classprior} devises a class-specific prior based on the band-pass filter responses and incorporates it into deblurring. In~\cite{featureprior}, two-directional two-dimensional PCA is used to extract feature-based sparse representation prior, thus mitigating the influence of blur and then achieving better image matching. As a result, a well-designed priori plays an important role in deblurring task. 

Unlike these empirical priors, the heterogeneous priors designed by us are directly related to each frame itself, which are essentially the explicit representation of input from multiple angles, and jointly serve as the input to the network with frames. It should be noted that, different from the conventional prior, the ``prior" in our work, on the one hand, refers to the inherent information of scenes, and on the other hand, it refers to the practice of using structure, motion, and contrast information to assist deblurring. In addition, as far as we know, this is also the first work to explicitly utilize priori information in video deblurring.  
\section{Proposed method}
\begin{figure*}[!t]
	\begin{center}
		\includegraphics[width=1.0\linewidth]{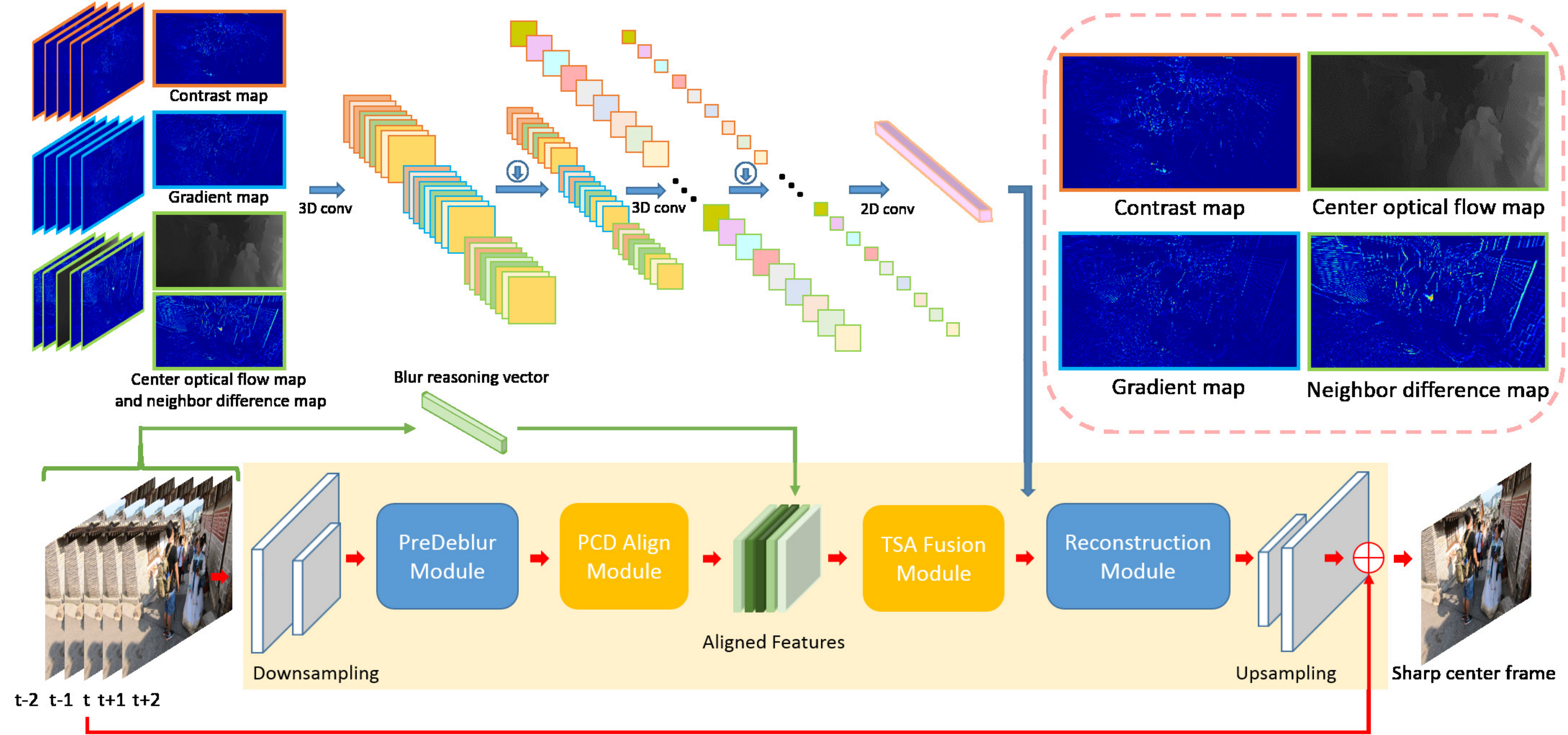} 
	\end{center}
	\caption{The diagram of our proposed PRiOr-enlightened and MOTION-robust video deblurring (PROMOTION) method. The modules in the golden background area are consistent with those in EDVR~\cite{edvr}. Recommend reading in color version.}
	\label{fig:framework}
\end{figure*}

The overall diagram of our proposed PRiOr-enlightened and MOTION-robust video deblurring (PROMOTION) method is presented in Fig.~\ref{fig:framework}. Given 5 consecutive input frames $I_{[t-2:t+2]}$ where $t$ is the center frame's number in the sequence, we denote the center frame as $I_{t}$ and the other frames as neighboring frames. The aim of video deblurring is to restore a sharp center frame $\hat{O_{t}}$ which is close to its corresponding ground truth $O_{t}$. On the one hand, we explicitly calculate the heterogeneous prior information of the input frame stack and encode it with 3D convolution to obtain a prior feature map, which is then multiplied by the features output from the TSA fusion module. This is described in Sec.~\ref{section:3.1}. On the other hand, we design the blur reasoning vector to rectify the aligned feature and introduce the optical flow based attention information in the loss function, to increase the model's perception ability of uneven blur distribution in spatio-temporal dimension. These are described in Sec.~\ref{section:3.2} and Sec.~\ref{section:3.4} separately. Finally, in order to make the model learn deblurring patterns more flexibly, channel attention is introduced into the basic residual blocks, which is described in Sec.~\ref{section:3.3}.
\subsection{Heterogeneous prior information}\label{section:3.1}
As discussed earlier in the introduction, we use the contrast, structure and motion priors of scenes to improve the model's structural fidelity and detail recovery abilities for low contrast and large optical flow regions.

\begin{figure*}[t]
	\begin{center}
		\begin{minipage}[t]{0.49\linewidth}
			\centering
			\includegraphics[height=5cm]{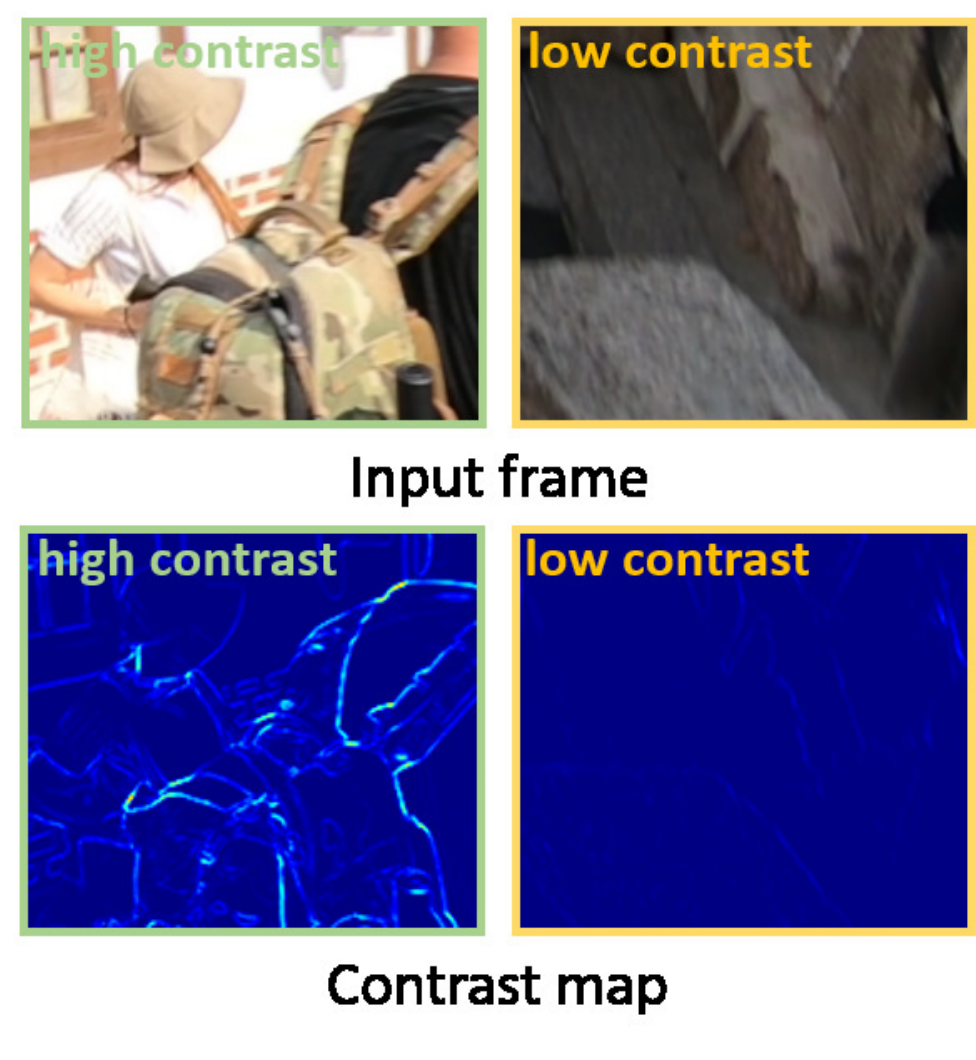} 
			\caption{Examples of different contrast.}
			\label{fig:contrast_cmp}
		\end{minipage}
		\space
		\begin{minipage}[t]{0.49\linewidth}
			\centering
			\includegraphics[height=5cm]{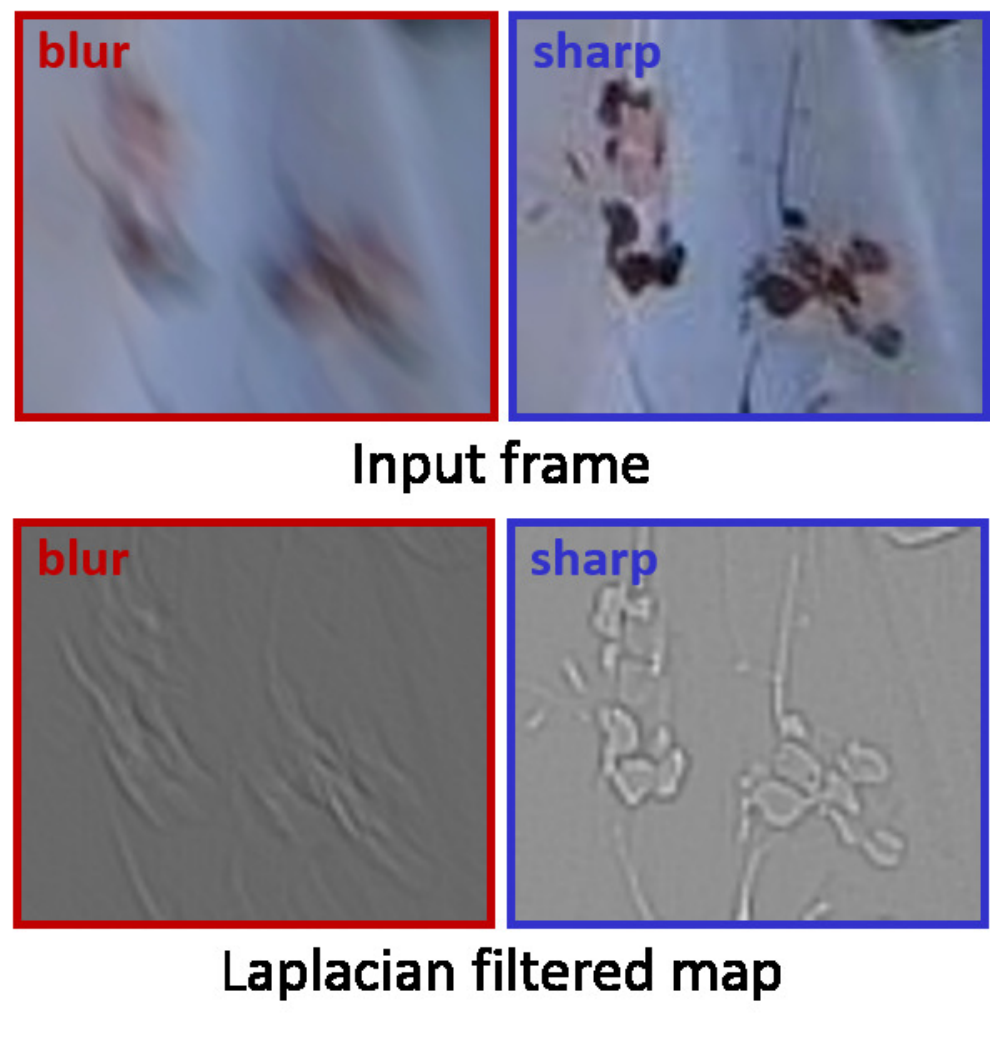}  
			\caption{Laplacian filtered maps.}
			\label{fig:lap_filter_cmp}
		\end{minipage}
	\end{center}	
\end{figure*}

\noindent\textbf{Contrast group.}
For each frame $I_{i}$ $(i\in[t-2,t+2])$ in the input stack, we calculate its contrast map $G_{i}^c$ as follows:
\begin{equation}
G_{i}^{c'}(p,q)=\frac{1}{4}\sum\limits_{(\hat{p},\hat{q})\in N_4(p,q)}(G_i(p,q)-G_i(\hat{p},\hat{q}))^2
\end{equation}
\begin{equation}
G_{i}^{c}=\frac{G_{i}^{c'}}{max(G_{i}^{c'})}
\end{equation}
where $G_{i}$ is the gray map of $I_{i}$, and $N_4(p,q)$ denotes the 4-neighborhood of pixel $(p,q)$. From the contrast map in Fig.~\ref{fig:contrast_cmp}, it can be seen that there are larger activation values for the high contrast regions and smaller activation values for the low ones. 

%

\noindent\textbf{Gradient group.}
Similarly, we calculate the gradient map $G_{i}^g$ for each input frame as
the equation (\ref{eq3}) given, and use it to increase the sensitivity of the model to structure information.
\begin{equation}\label{eq3}
G_{i}^g(p,q)=[G_{i}(p,q)-G_{i}(p+1,q)]+[G_{i}(p,q)-G_{i}(p,q+1)]
\end{equation}
As the pink box shown in the upper right corner of Fig.~\ref{fig:framework}, gradient map can highlight structural information in a scene, such as regular lines on the wall. This allows the model to have better structural fidelity.

\noindent\textbf{Motion group.}
Instead of using all the optical flow maps of input stack, we only estimate the optical flow map of center frame as baseline by using~\cite{flownet2.0}, while for the neighbor frames, we use the difference maps between center frame and each neighbor frame to represent the relative motion information. This is to save the calculation time of optical flow.

Then 3D group convolutions are used to encode the spatio-temporal prior information more efficiently. The detail parameters of the heterogeneous prior encoding network are presented in Table~\ref{table:3Dconv_para}.
\setlength{\tabcolsep}{4pt}
\begin{table}[]
	\caption{Parameters of the heterogeneous prior encoding network. Note stride is the same in row, column and depth directions.}
	\begin{center}
		\renewcommand\arraystretch{0.9}
		\resizebox{0.8\columnwidth}{!}{ 
			\begin{tabular}{c||ccccc}
				\hline
				Layer                        & In\_channel & Out\_channel & Kernel Size & Stride & Group\\ \hline
				\multicolumn{1}{c||}{3D Conv1}   & 3          & 9          & 3*5*5         & 1  & 3    \\
				\multicolumn{1}{c||}{MaxPool} & 9          & 9           & 2*2         & 2    & - \\
				\multicolumn{1}{c||}{3D Conv2}   & 9          & 27          & 3*5*5         & 1  & 9    \\
				\multicolumn{1}{c||}{MaxPool}   & 27         & 27          & 2*2         & 2    & - \\
				\multicolumn{1}{c||}{2D Conv}   & 27         & 128           & 1*1         & 1  & - \\ \hline
		\end{tabular}}
	\end{center}
	\label{table:3Dconv_para}
\end{table}
\setlength{\tabcolsep}{1.4pt}

\subsection{Blur reasoning vector}\label{section:3.2}
In order to enable the model to place corresponding emphasis on frames with different degrees of blur in temporal dimension, blur reasoning vector is designed to explicitly tell the model which frames it should pay more attention to. 

Specifically, we first filter each frame with the Laplacian operator of size $3*3$, as shown in Fig.~\ref{fig:lap_filter_cmp}. It is not difficult to find that the blurry image on the left has a smooth filtered map, while the clear one on the right has a sharper filtered result. Then, we calculate the variance of these filtered maps $var(Lap(G_i))$ to reflect the blur degree of each frame. The more blurry the frame, the smaller its variance. Therefore, the blur reasoning vector can be represented by:
\begin{equation}
V_{t}^{blur}=\frac{5\cdot[var(Lap(G_i))^{-1}]}{\sum_{i=t-2}^{t+2}var(Lap(G_i))^{-1}},\quad i\in[t-2,t+2]
\end{equation}
The size of $V_{t}^{blur}$ is 5, whose each dimension indicates the importance of the $i^{th}$ frame. Finally, this vector is multiplied by the aligned feature in depth and we obtain the refined aligned feature.

\subsection{Channel attention enhanced residual block}\label{section:3.3}
\begin{figure}[t]
	\begin{center}
		\includegraphics[width=1.0\columnwidth]{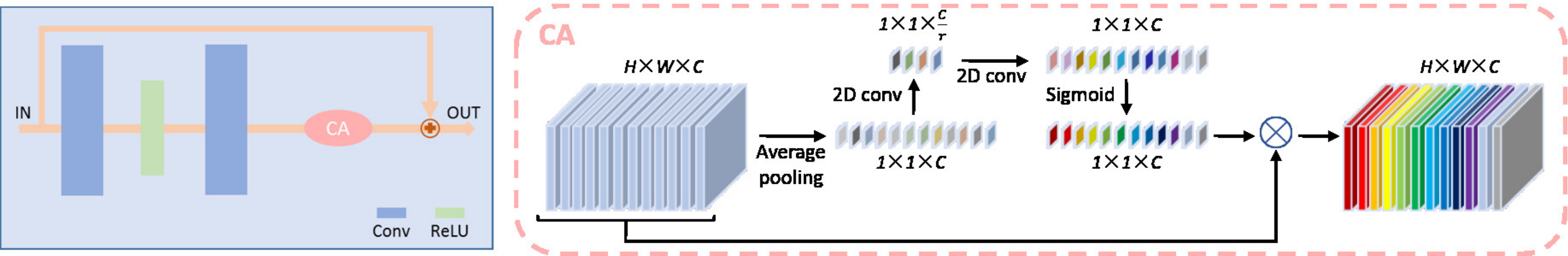} 
	\end{center}
	\caption{Illustration of channel attention (CA) used in the residual basic block.}
	\label{fig:channel_attention}
\end{figure}

Inspired by the success of channel attention in super resolution~\cite{rcan}, in the residual basic block, we also introduce this mechanism to adaptively enhance and organize the necessary patterns related to deblurring. As shown in Fig.~\ref{fig:channel_attention}, on the one hand, the dimensions of channel are first compressed and then restored, therefore the effective information can be amplified. On the other hand, by adaptively learning the importance of each channel, the model can express deblurring patterns more flexibly.

\subsection{Dual loss function}\label{section:3.4}
In order to increase the naturalness of deblurred videos, we constrain the training process from two complementary levels, namely pixel level and perceptual level. 

At the pixel level, we believe that the blurred areas are often the areas suffering from motion or change, while these are exactly the focus of people when watching videos. For example, in the hand pose estimation scenario, compared with the still background, people usually pay more attention to the movement of hands. As a result, the optical flow information of the current frame is used as an indicator of attention, to increase the model's sense ability for non-uniform blur in spatio domain, which complements the blur reasoning vector perceiving in temporal domain. Therefore, the pixel-wise loss can be represented as below:
\begin{equation}
L_{cb}=\frac{1}{HW}\sum\sqrt{((\hat{O_t}-O_t)\otimes(1+w_{att}))^2+\epsilon}
\end{equation}
where $\otimes$ denotes element-wise multiplication and $\epsilon$ is $10^{-6}$. $w_{att}$ is the normalized optical flow map. It should be noted that, the optical flow is first converted to an RGB image according to the color encoding system, then it is converted to a grayscale image. Therefore, the flow map mentioned is $\in R^{H*W}$. $H$ and $W$ are the height and width of one frame.

At the perceptual level, in order to describe the subjective visual differences between deblurred frames and ground truth, we use neural network-based perceptual similarity to represent such high-level distance~\cite{perceptualsimilarity} as below:
\begin{equation}
L_{ps}=f_{ps}(\hat{O_t},O_t)
\end{equation}
where $f_{ps}$ is the perceptual similarity network, whose weights have been pretrained on a large-scale subjective dataset and output is a score ranging from 0 to 1.

Therefore, the overall loss function is:
\begin{equation}
L=L_{cb}+\lambda L_{ps}
\end{equation}
Here $\lambda$ is a balance factor adjusting the relative importance of pixel-level and perceptual-level losses. We empirically set it as $0.1$. In this way, the model can not only handle non-uniform blur, but also guarantee the subjective quality of deblurred videos.

\section{Experiments}
In this section, we first verify the motion robustness of our model in the global and local blurry scenarios. For global blur, two well-known video deblurred datasets for natural scenes, involving camera shake and object moving, are used for evaluation in Sec.~\ref{section:4.1}. For local blur, in Sec.~\ref{section:4.2}, we consider the specific application scenario, hand pose estimation, and construct the first video deblurring dataset for hand pose estimation. In this case, the estimation accuracy is also used as one of the measurement of deblurring effect. Then we conduct the ablation studies to prove the validity of each module in Sec.~\ref{section:4.3}. Finally, the effectiveness of handling challenging blurs, and the complexity and parameters of the model are further discussed in Sec.~\ref{section:4.4}.

\subsection{Global blur}\label{section:4.1}
\setlength{\tabcolsep}{8pt}
\begin{table*}[!t]
	\centering
	\caption{Performance comparison under the standard division of REDS dataset.}
	\resizebox{0.58\linewidth}{!}{ 
		\begin{threeparttable}
			\begin{tabular}{c||c|c|c}
				\hline 
				Metric                      & Fan's~\cite{anempirical} & Sim's~\cite{rdurud} & PROMOTION   \\ \hline
				\multicolumn{1}{c||}{PSNR}   & 34.17   & 33.38   & \textbf{34.25}  \\ \hline
				\multicolumn{1}{c||}{SSIM}   & 0.9345  & -       & \textbf{0.9536} \\ \hline
				\multicolumn{1}{c||}{Note}   & $2^{nd}$& $3^{rd}$ & -      \\ \hline
			\end{tabular}
			\begin{tablenotes}
				\scriptsize 
				\item[*] where~\cite{anempirical} uses 8x self-ensemble. 
			\end{tablenotes}
	\end{threeparttable}}
	\label{table:reds_cmp1}
\end{table*}
\setlength{\tabcolsep}{1.4pt}

\begin{table*}[!t]
	\centering
	\caption{Performance comparison under the same dataset division of REDS dataset with training EDVR.}
	\resizebox{\linewidth}{!}{ 
		\begin{tabular}{c||c|c|c|c|c|c}
			\hline
			Metric                      & DeblurGAN~\cite{deblurgan} & Nah's~\cite{Nah_2017_CVPR} & SRN~\cite{Tao_2018_CVPR} & DeBlurNet~\cite{deblurnet} & EDVR~\cite{edvr} & PROMOTION \\ \hline
			\multicolumn{1}{c||}{PSNR}   & 24.09   & 26.16   & 26.98 & 26.55 & 34.80 & \textbf{35.10} \\ \hline
			\multicolumn{1}{c||}{SSIM}   & 0.7482  & 0.8249  & 0.8141  & 0.8066  & 0.9487  & \textbf{0.9565}  \\ \hline
	\end{tabular}}
	\label{table:reds_cmp2}
\end{table*}

\noindent\textbf{REDS dataset.}
This dataset used in the NTIRE19 challenge includes 300 videos, where 240 videos are for training, 30 videos are for validation and the remaining 30 videos are for testing~\cite{ntire}. However, only the training and validation sets (total 270 videos) are available online right now. For more comprehensive evaluation, we consider two kinds of dataset division. One is the standard division (240 training videos and 30 validation videos for testing), and the other one follows the same setting as EDVR, which wins the championship in the challenge, for fair comparison (266 training videos and 4 specific videos for testing)~\cite{edvr}. Each video has 100 frames. And their resolution is 720*1280 and frame rate is 24 fps.

Under the standard setting, we compare our method with the $2^{nd}$ and $3^{rd}$ methods in the challenge~\cite{anempirical,rdurud}, as the average results given in Table~\ref{table:reds_cmp1}. Even though~\cite{anempirical} has used eight times self-ensemble, our method still achieves the best performance without using any data augmentation trick.

\begin{figure*}[!t]
	\begin{center}
		\includegraphics[width=1.0\linewidth]{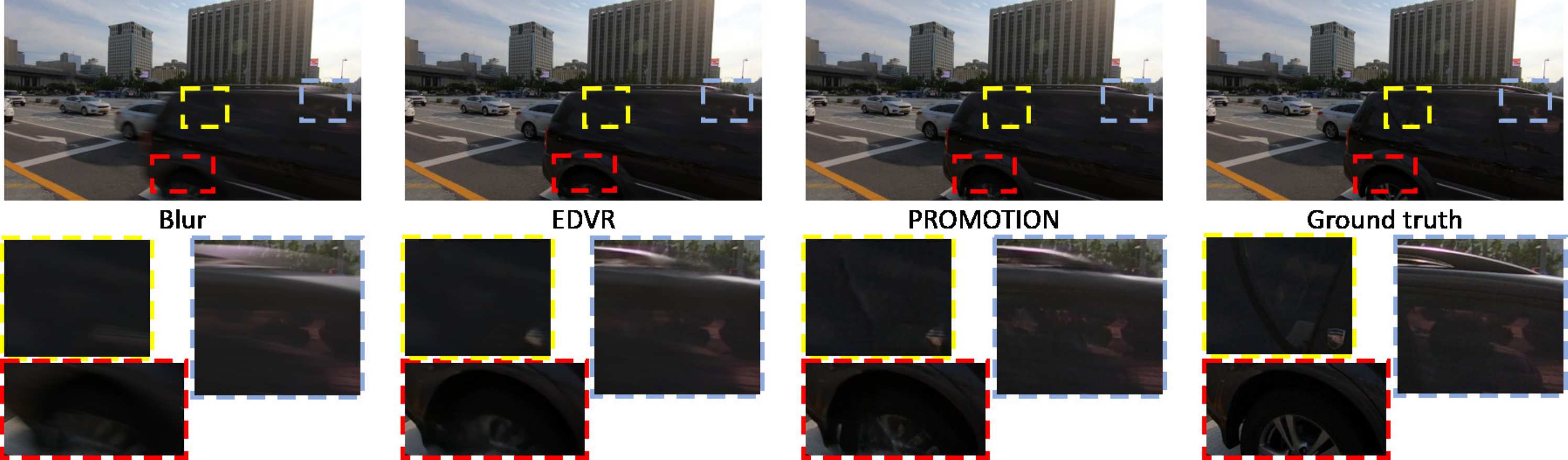} 
	\end{center}
	\caption{Qualitative comparison on \textbf{REDS} dataset. Please zoom in for best view.}
	\label{fig:reds_2_cmp}
\end{figure*}

Under the second setting, as presented in Table~\ref{table:reds_cmp2}, compared with the original EDVR and other four kinds of deblurring methods, namely  DeblurGAN~\cite{deblurgan}, Nah's~\cite{Nah_2017_CVPR}, SRN~\cite{Tao_2018_CVPR}, and DeBlurNet~\cite{deblurnet}, our model continues obtaining the best performance both in terms of PSNR and SSIM. Let us more intuitively analyze the performance gap with EDVR. As the tire, window and roof of the black car shown in the Fig.~\ref{fig:reds_2_cmp}, our method can restore more details for these challenging low-contrast areas. 

\noindent\textbf{GoPro dataset.}
To further demonstrate the robustness of our model, we also test it on another video deblurring dataset for natural scenes named GoPro~\cite{Nah_2017_CVPR}. 22 training sequences and 11 testing sequences are included in it. Each sequence has unequal lengths, but the resolution is the same one 720*1280. It should be noted that the dataset provides two kinds of blurry images, including gamma corrected and linear CRF versions.

\begin{figure*}[!t]
	\begin{center}
		\includegraphics[width=1.0\linewidth]{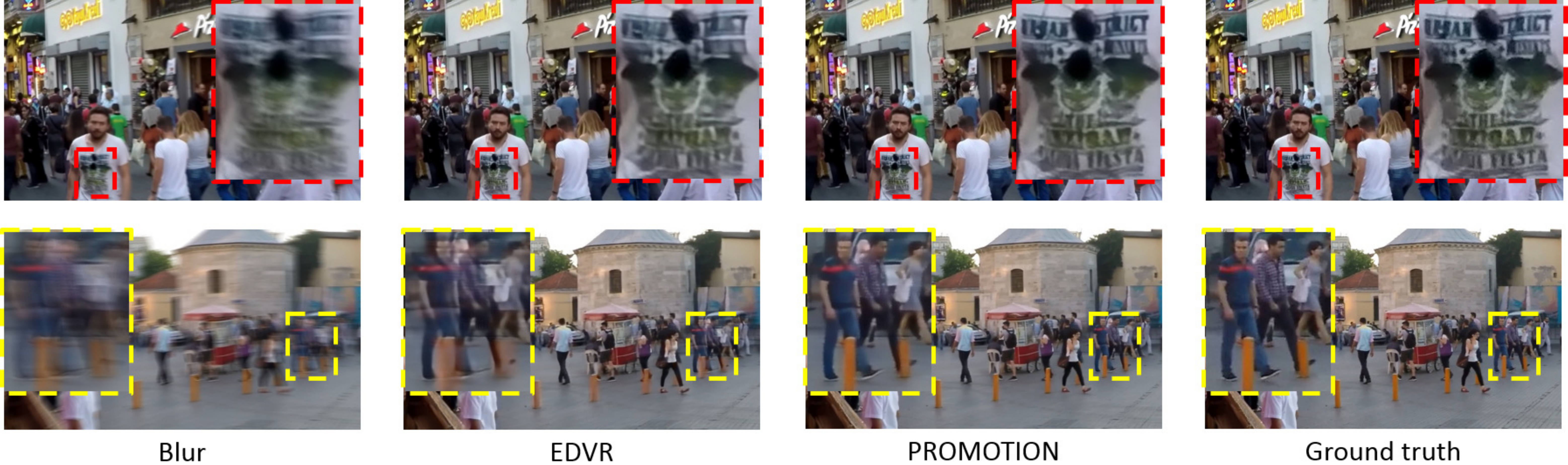} 
	\end{center}
	\caption{Qualitative comparison on \textbf{GoPro} dataset. Please zoom in for best view.}
	\label{fig:gopro_cmp}
\end{figure*}

For the evaluation of EDVR and our model on GoPro dataset, we finetune the models pretrained on REDS dataset. As Table~\ref{table:gopro_cmp} presented, where all the models are tested on the linear CRF version, we compare with both video deblurring~\cite{recurrentZhang_2018_CVPR,rdurud,edvr} and image debluring methods~\cite{Nah_2017_CVPR,deblurganv2,Tao_2018_CVPR}. Our model outperforms the state-of-the-art methods by a large margin, both in terms of video or image deblurring. Compared with the improvement on REDS dataset, we attribute this to both the effectiveness of prior information and the lower image quality of GoPro dataset. First, using the prior information can alleviate the model's dependence on datasets and assist the model in obtaining information about the current data distribution more directly and comprehensively. Second, according to the shielding effect, we can infer that on the basis of low quality, the improvement is obvious, while on the basis of high quality, the difference is less conspicuous. Qualitative results are given in Fig.~\ref{fig:gopro_cmp}. Although different pedestrians in the scene have various motion modes, it is easy to see that our method can better deal with this non-uniform blur and achieve the deblurring effect closer to ground truth.

\begin{table*}[!t]
	\caption{Performance comparison on GoPro dataset (\textit{linear CRF ver.}).}
	\begin{center}
		\resizebox{\linewidth}{!}{ 
			\begin{tabular}{c||c|c|c|c|c|c|c}
				\hline
				Metric                      & Nah's~\cite{Nah_2017_CVPR} & DeblurGAN-v2~\cite{deblurganv2} & SRN~\cite{Tao_2018_CVPR} & Zhang's~\cite{recurrentZhang_2018_CVPR} &Sim's~\cite{rdurud} & EDVR~\cite{edvr} & PROMOTION \\ \hline
				\multicolumn{1}{c||}{PSNR}   & 28.62   & 29.55   & 30.26 & 29.19 & 31.34 & 30.20 &\textbf{33.25} \\ \hline
				\multicolumn{1}{c||}{SSIM}   & 0.9094  & 0.9340  & 0.9342  & 0.9306  & 0.9474 & 0.9109  & \textbf{0.9481}  \\  \hline
				\multicolumn{1}{c||}{Note}   &\multicolumn{3}{c|}{Image deblurring}  & \multicolumn{4}{c}{Video deblurring}  \\ \hline
		\end{tabular}}
	\end{center}
	\label{table:gopro_cmp}
\end{table*}

\subsection{Local blur}\label{section:4.2}
In order to show the generalization of the model, we consider the specific local blurry downstream task, hand pose estimation. First, a video deblurring dataset for hand pose estimation is synthesized. Then we test the improvement of estimation accuracy after deblurring to indirectly reflect the deblurring effect.

\begin{figure}[!t]
	\begin{center}
		\includegraphics[width=1.0\linewidth]{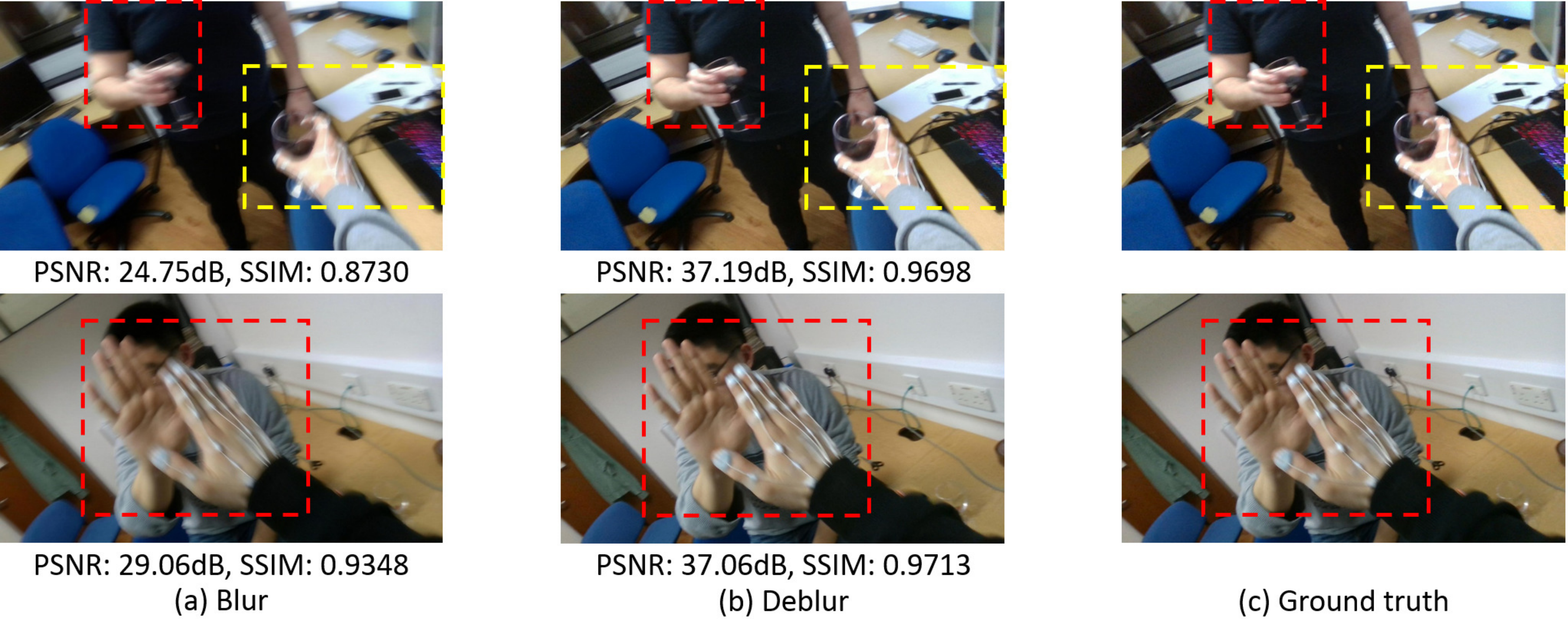} 
	\end{center}
	\caption{Deblurring performance on the synthesized video deblurring dataset for hand pose estimation. Please zoom in for best view.}
	\label{fig:handpose_cmp1}
\end{figure}

Based on the existing hand pose estimation dataset named First-Person Hand Action Benchmark~\cite{firstperson}, we synthesize blur for these sequences. This dataset provides RGB-D frames and their corresponding hand pose labels. All the sequences have a frame rate of 30 fps and resolution of 1080*1920. Then we follow the same blur synthesis process as REDS dataset~\cite{ntire}. The frames are interpolated to virtually increase the frame rate up to 1920 fps by~\cite{frameinterpolation}, which makes the averaged frames exhibit smooth and realistic blurs without spikes and step artifacts. Finally, the virtual frames are averaged in the signal space to mimic camera imaging pipeline as below~\cite{Nah_2017_CVPR}:
\begin{equation}
B=g(\frac{1}{m}\sum_{i=0}^{m-1}S[i])
\end{equation}
Here $g(x)=x^{\frac{1}{\gamma}}$ is an approximated camera response function, where $\gamma$ is usually set as 2.2. $B$ represents the blur frame, and $S[i]$ is the latent sharp frame which can be calculated by our observed sharp frame $\hat{S[i]}$: $S[i]=g^{-1}(\hat{S[i]})$.
The resulting blurry videos have a frame rate of 30 fps. The synthesized blur effect is shown in Fig.~\ref{fig:handpose_cmp1} (a) and (c).

\begin{figure}[!t]
	\begin{center}
		\includegraphics[width=\linewidth]{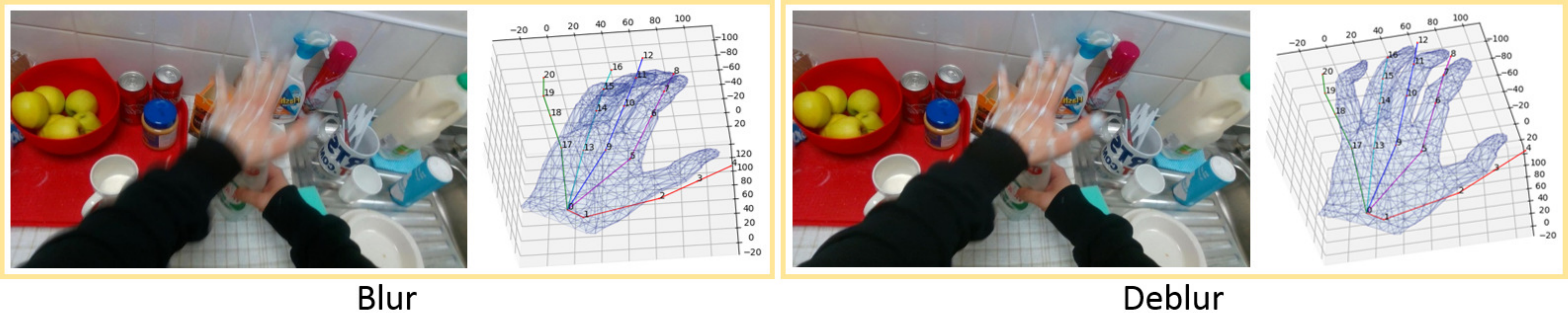} 
	\end{center}
	\caption{Performance of hand pose estimation before and after deblurring. The improvement of RMSE and ABSE of joint location estimatioin are 11.9015 and 9.4662.}
	\label{fig:handpose_cmp2}
\end{figure}

We finetune the model pretrained on REDS dataset, and use the state-of-the-art hand pose estimation method~\cite{learningjoint} to measure the accuracy. Here we calculate the RMSE and absolute error (ABSE) between the locations of the estimated joints and ground truth as metrics. As expected, in the Table~\ref{table:hand_cmp}, all the indicators improve after deblurring. Visualized results are given in Fig.~\ref{fig:handpose_cmp1} (b) and Fig.~\ref{fig:handpose_cmp2}. In Fig.~\ref{fig:handpose_cmp2}, skeleton represents ground truth, and mesh represents the estimated results. It is not difficult to see that for the deblurred frame, skeleton and mesh have a better overlap. This indicates our work can be used as an input preprocessing method to bring gains to downstream machine tasks.


\begin{table}[!t]
	\begin{minipage}[!t]{0.48\columnwidth}
		\renewcommand{\arraystretch}{1.37}
		\caption{Performance improvement for hand pose estimation after deblurring.}
		\begin{center}
			\setlength{\tabcolsep}{8pt}
			\resizebox{\linewidth}{!}{ 
				\begin{threeparttable}
					\begin{tabular}{c||c|c|c|c}
						\hline
						& PSNR & SSIM & RMSE & ABSE   \\ \hline
						\multicolumn{1}{c||}{$\bigtriangleup$}   & 6.0171 & 0.0287 & 0.3110 & 0.2653      \\ \hline
					\end{tabular}
					\begin{tablenotes}
						\scriptsize 
						\item[*] where $\bigtriangleup$ means the improvement of values.  
					\end{tablenotes}
			\end{threeparttable}}
			\setlength{\tabcolsep}{1.4pt}
		\end{center}
		\label{table:hand_cmp}
	\end{minipage}
	\quad
	\begin{minipage}[!t]{0.48\columnwidth}
		\caption{Comparison in complexity and params of models.}
		\begin{center}
			\resizebox{1.0\linewidth}{!}{ 
				\begin{threeparttable}
					\begin{tabular}{c||c|c|c}
						\hline
						& EDVR~\cite{edvr} & Ours & $\bigtriangleup$ \\ \hline
						\multicolumn{1}{c||}{FLOPs (GMac)}   & 194.78 & 195.02 & \textbf{0.24 (0.12$\%$)} \\ \hline 
						\multicolumn{1}{c||}{Params (M)}   & 23.60 & 23.72 & \textbf{0.12 (0.51$\%$)}\\ \hline
					\end{tabular}
					\begin{tablenotes}
						\scriptsize 
						\item[*] FLOPs are for images of size $256*256$.  
					\end{tablenotes}
			\end{threeparttable}}
		\end{center}
		\label{table:complexity and params}
	\end{minipage}
	\vspace{-0.1cm}
\end{table}

\begin{table*}[!t]
	\caption{Performance of ablation studies on GoPro dataset (\textit{gamma corrected ver.}).}
	\begin{center}
		\resizebox{\linewidth}{!}{ 
			\begin{threeparttable}
				\begin{tabular}{c||c|c|c|c|c|c|c|c|c|c}
					\hline
					& EDVR~\cite{edvr} & Ab1 & Ab1-i & Ab1-s & Ab1-m & Ab2 & Ab3 & Ab4 & Ab5 & All  \\ \hline
					\multicolumn{1}{c||}{PSNR}   & 29.76 & 31.57 & 31.74 & 31.74 & 31.83 & 33.14 & 30.23 & 33.11 & 33.28 & \textbf{33.28}      \\ \hline
					\multicolumn{1}{c||}{SSIM}   & 0.9041 & 0.9264 & 0.9328 & 0.9320 & 0.9326 & 0.9461 & 0.9100 & 0.9457 & 0.9472 & \textbf{0.9477} \\ \hline
				\end{tabular}
				\begin{tablenotes}
					\scriptsize 
					\item[*] Ab1: w/o heterogeneous prior information, Ab1-[i/s/m]: w/o only [contrast/structure/motion] prior information, Ab2: w/o blur reasoning vector, Ab3: w/o channel attention, Ab4: w/o optical flow based attention in the Charbonnier loss, Ab5: w/o perceptual-level loss.  
				\end{tablenotes}
		\end{threeparttable}} 
	\end{center}
	\label{table:abalation study}
\end{table*}

\subsection{Ablation studies}\label{section:4.3}
\begin{figure}[!t]
	\begin{center}
		\includegraphics[width=1.0\columnwidth]{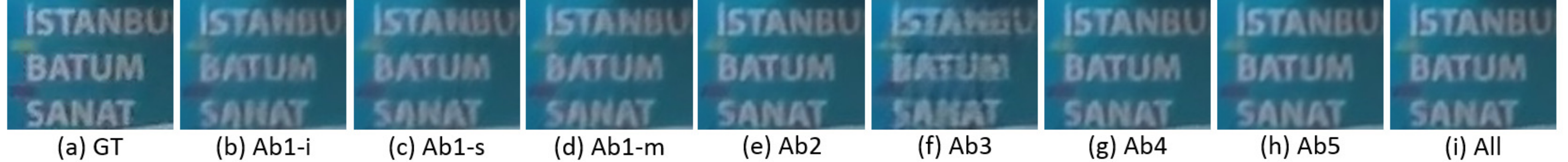} 
	\end{center}
	\caption{Visulization of the ablation study. Please zoom in for best view.}
	\label{fig:ab_visualization}
\end{figure}

Instead of using linear CRF image subset, we choose the gamma corrected images of GoPro dataset as the evaluation data for ablation study, to further demonstrate the model's robustness. We test the models of removing each module separately to prove its effectiveness. According to Table~\ref{table:abalation study}, we can find that heterogeneous prior information and channel attention play more significant roles. This indirectly shows that the conventional model does not sufficiently mine data information, because neither the model structure nor the loss function particularly consider challenging blurs. However, by explicitly emphasizing this part information in input, we can achieve the promising results. What needs to be explained is the effect of the blur reasoning vector seems to be diminished due to the flat temporal blur variotion. In addition, we visualize the deblurring results of each ablation setting in Fig.~\ref{fig:ab_visualization}. It can be seen that the use of prior information significantly enhances the visual subjective effect.

\subsection{Analysis}\label{section:4.4}
\begin{figure}[!t]
	\begin{center}
		\includegraphics[width=1.0\columnwidth]{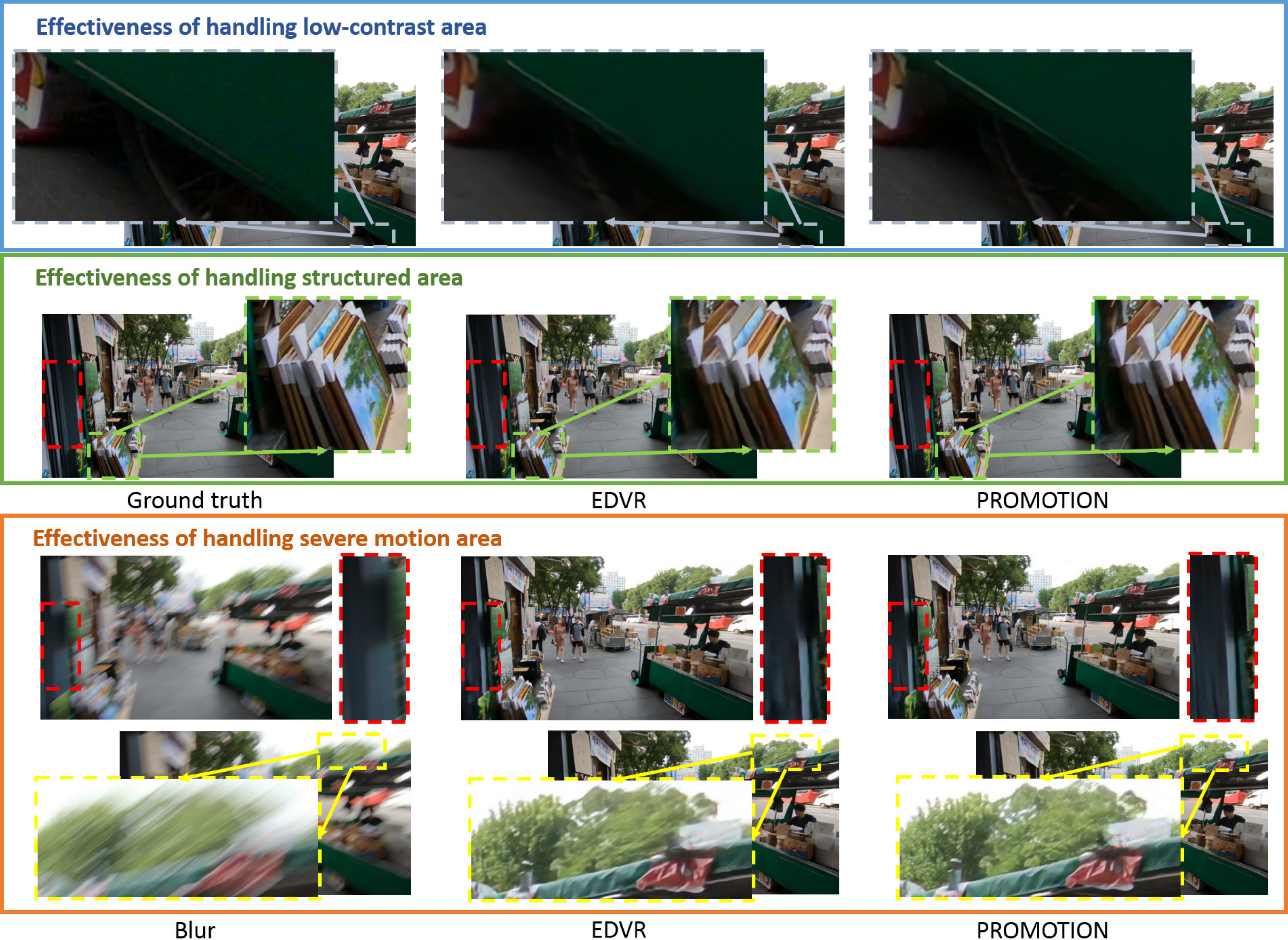} 
	\end{center}
	\caption{Illustration of the effectiveness of handling challenging blurs.}
	\label{fig:prior_analyze}
\end{figure}

\noindent\textbf{Effectiveness of handling challenging blurs.}
Corresponding to our work's motivation, next we visualize the performance of deblurring challenging blurs. As the blue block shown in Fig.~\ref{fig:prior_analyze}, our method has a better recovery capability for the low-contrast tire than EDVR. And in the middle green block, we can see that our method has better structual fidelity of the photo frames, which indirectly reflects the effective constraint of gradient prior. For the close shot areas with severe blur, as the orange block presented, our method is able to recover more details thanks to the consolidation of motion prior.

\noindent\textbf{Complexity and params of the model.}
Although we have proposed a series of effective and plug-and-play
designs for deblurring, and achieved the state-of-the-art performance, while as presented in the Table~\ref{table:complexity and  params}, the increase in FLOPs and params is almost negligible.

\section{Conclusion}
To better handle challenging blurs, we first introduce prior information in video deblurring, and propose a PRiOr-enlightened and MOTION-robust video deblurring (PROMOTION) model. Specifically, 3D group convolutions are used to better encode heterogeneous priors, including contrast, structure, and motion information of scenes, which are proven to be related with deblurring. Then, blur reasoning vector and optical-flow based attention prior are used to increase the model's ability to recognize spatio-temporal non-uniform blur. Experimental results on two globally blurred datasets show our method can achieve the state-of-the-art performance. In addition, for downstream applications suffering from local blur, such as hand pose estimation, we also demonstrate that our method can bring performance gains to the task by preprocessing source data.

We believe our work can shed light not only on video deblurring in respect of utilizing the inherent information of scenes, but also on how to measure video deblurring in respect of borrowing downstream tasks. In the future, the prior encoding branch, as a portable module, will be integrated into more existing video deblurring methods and other enhancement tasks to achieve more visually pleasing restoration performance. On the other hand, more downstream tasks will be considered to indirectly measure the deblurring quality, thereby optimizing more machine-based applications.

\bibliography{promotion_pr}

\begin{thebibliography}{10}
\expandafter\ifx\csname url\endcsname\relax
  \def\url#1{\texttt{#1}}\fi
\expandafter\ifx\csname urlprefix\endcsname\relax\def\urlprefix{URL }\fi
\expandafter\ifx\csname href\endcsname\relax
  \def\href#1#2{#2} \def\path#1{#1}\fi

\bibitem{handpose1}
R.~Li, Z.~Liu, J.~Tan,
  \href{http://www.sciencedirect.com/science/article/pii/S0031320319301724}{A
  survey on 3d hand pose estimation: Cameras, methods, and datasets}, Pattern
  Recognition 93 (2019) 251 -- 272.
\newblock \href
  {http://dx.doi.org/https://doi.org/10.1016/j.patcog.2019.04.026}
  {\path{doi:https://doi.org/10.1016/j.patcog.2019.04.026}}.
\newline\urlprefix\url{http://www.sciencedirect.com/science/article/pii/S0031320319301724}

\bibitem{learningjoint}
Y.~Hasson, G.~Varol, D.~Tzionas, I.~Kalevatykh, M.~J. Black, I.~Laptev,
  C.~Schmid, Learning joint reconstruction of hands and manipulated objects,
  in: The IEEE Conference on Computer Vision and Pattern Recognition (CVPR),
  2019.

\bibitem{handpose2}
T.-Y. Chen, P.-W. Ting, M.-Y. Wu, L.-C. Fu,
  \href{http://www.sciencedirect.com/science/article/pii/S0031320318300839}{Learning
  a deep network with spherical part model for 3d hand pose estimation},
  Pattern Recognition 80 (2018) 1 -- 20.
\newblock \href
  {http://dx.doi.org/https://doi.org/10.1016/j.patcog.2018.02.029}
  {\path{doi:https://doi.org/10.1016/j.patcog.2018.02.029}}.
\newline\urlprefix\url{http://www.sciencedirect.com/science/article/pii/S0031320318300839}

\bibitem{facerecog1}
X.~Wei, H.~Wang, B.~Scotney, H.~Wan,
  \href{http://www.sciencedirect.com/science/article/pii/S0031320319303152}{Minimum
  margin loss for deep face recognition}, Pattern Recognition 97 (2020) 107012.
\newblock \href
  {http://dx.doi.org/https://doi.org/10.1016/j.patcog.2019.107012}
  {\path{doi:https://doi.org/10.1016/j.patcog.2019.107012}}.
\newline\urlprefix\url{http://www.sciencedirect.com/science/article/pii/S0031320319303152}

\bibitem{facerecog2}
M.~He, J.~Zhang, S.~Shan, M.~Kan, X.~Chen,
  \href{http://www.sciencedirect.com/science/article/pii/S0031320319304145}{Deformable
  face net for pose invariant face recognition}, Pattern Recognition 100 (2020)
  107113.
\newblock \href
  {http://dx.doi.org/https://doi.org/10.1016/j.patcog.2019.107113}
  {\path{doi:https://doi.org/10.1016/j.patcog.2019.107113}}.
\newline\urlprefix\url{http://www.sciencedirect.com/science/article/pii/S0031320319304145}

\bibitem{reid3}
A.~Serbetci, Y.~S. Akgul,
  \href{http://www.sciencedirect.com/science/article/pii/S0031320320301229}{End-to-end
  training of cnn ensembles for person re-identification}, Pattern Recognition
  104 (2020) 107319.
\newblock \href
  {http://dx.doi.org/https://doi.org/10.1016/j.patcog.2020.107319}
  {\path{doi:https://doi.org/10.1016/j.patcog.2020.107319}}.
\newline\urlprefix\url{http://www.sciencedirect.com/science/article/pii/S0031320320301229}

\bibitem{reid4}
H.~Luo, W.~Jiang, X.~Zhang, X.~Fan, J.~Qian, C.~Zhang,
  \href{http://www.sciencedirect.com/science/article/pii/S0031320319302031}{Alignedreid++:
  Dynamically matching local information for person re-identification}, Pattern
  Recognition 94 (2019) 53 -- 61.
\newblock \href
  {http://dx.doi.org/https://doi.org/10.1016/j.patcog.2019.05.028}
  {\path{doi:https://doi.org/10.1016/j.patcog.2019.05.028}}.
\newline\urlprefix\url{http://www.sciencedirect.com/science/article/pii/S0031320319302031}

\bibitem{Nah_2017_CVPR}
S.~Nah, T.~Hyun~Kim, K.~Mu~Lee, Deep multi-scale convolutional neural network
  for dynamic scene deblurring, in: The IEEE Conference on Computer Vision and
  Pattern Recognition (CVPR), 2017.

\bibitem{Tao_2018_CVPR}
X.~Tao, H.~Gao, X.~Shen, J.~Wang, J.~Jia, Scale-recurrent network for deep
  image deblurring, in: The IEEE Conference on Computer Vision and Pattern
  Recognition (CVPR), 2018.

\bibitem{multiscaleprior}
Y.~{Bai}, H.~{Jia}, M.~{Jiang}, X.~{Liu}, X.~{Xie}, W.~{Gao}, Single image
  blind deblurring using multi-scale latent structure prior, IEEE Transactions
  on Circuits and Systems for Video Technology (2019) 1--1\href
  {http://dx.doi.org/10.1109/TCSVT.2019.2919159}
  {\path{doi:10.1109/TCSVT.2019.2919159}}.

\bibitem{deblurganv2}
O.~Kupyn, T.~Martyniuk, J.~Wu, Z.~Wang, Deblurgan-v2: Deblurring
  (orders-of-magnitude) faster and better (2019).
\newblock \href {http://arxiv.org/abs/1908.03826} {\path{arXiv:1908.03826}}.

\bibitem{edvr}
X.~Wang, K.~C. Chan, K.~Yu, C.~Dong, C.~Change~Loy, Edvr: Video restoration
  with enhanced deformable convolutional networks, in: The IEEE Conference on
  Computer Vision and Pattern Recognition (CVPR) Workshops, 2019.

\bibitem{deblurnet}
S.~Su, M.~Delbracio, J.~Wang, G.~Sapiro, W.~Heidrich, O.~Wang, Deep video
  deblurring for hand-held cameras, in: The IEEE Conference on Computer Vision
  and Pattern Recognition (CVPR), 2017.

\bibitem{recurrentKim_2017_ICCV}
T.~Hyun~Kim, K.~Mu~Lee, B.~Scholkopf, M.~Hirsch, Online video deblurring via
  dynamic temporal blending network, in: The IEEE International Conference on
  Computer Vision (ICCV), 2017.

\bibitem{recurrentZhang_2018_CVPR}
J.~Zhang, J.~Pan, J.~Ren, Y.~Song, L.~Bao, R.~W. Lau, M.-H. Yang, Dynamic scene
  deblurring using spatially variant recurrent neural networks, in: The IEEE
  Conference on Computer Vision and Pattern Recognition (CVPR), 2018.

\bibitem{recurrentNah_2019_CVPR}
S.~Nah, S.~Son, K.~M. Lee, Recurrent neural networks with intra-frame
  iterations for video deblurring, in: The IEEE Conference on Computer Vision
  and Pattern Recognition (CVPR), 2019.

\bibitem{adversarialSTlearning}
K.~{Zhang}, W.~{Luo}, Y.~{Zhong}, L.~{Ma}, W.~{Liu}, H.~{Li}, Adversarial
  spatio-temporal learning for video deblurring, IEEE Transactions on Image
  Processing 28~(1) (2019) 291--301.
\newblock \href {http://dx.doi.org/10.1109/TIP.2018.2867733}
  {\path{doi:10.1109/TIP.2018.2867733}}.

\bibitem{anempirical}
Y.~Fan, J.~Yu, D.~Liu, T.~S. Huang, An empirical investigation of efficient
  spatio-temporal modeling in video restoration, in: The IEEE Conference on
  Computer Vision and Pattern Recognition (CVPR) Workshops, 2019.

\bibitem{ntire}
S.~Nah, R.~Timofte, S.~Baik, S.~Hong, G.~Moon, S.~Son, K.~Mu~Lee, Ntire 2019
  challenge on video deblurring: Methods and results, in: The IEEE Conference
  on Computer Vision and Pattern Recognition (CVPR) Workshops, 2019.

\bibitem{charbonnierloss}
P.~{Charbonnier}, L.~{Blanc-Feraud}, G.~{Aubert}, M.~{Barlaud}, Two
  deterministic half-quadratic regularization algorithms for computed imaging,
  in: Proceedings of 1st International Conference on Image Processing, Vol.~2,
  1994, pp. 168--172 vol.2.
\newblock \href {http://dx.doi.org/10.1109/ICIP.1994.413553}
  {\path{doi:10.1109/ICIP.1994.413553}}.

\bibitem{tra_deblur1}
L.~Zhang, Optical flow in the presence of spatially-varying motion blur, in:
  Proceedings of the 2012 IEEE Conference on Computer Vision and Pattern
  Recognition (CVPR), CVPR ’12, IEEE Computer Society, USA, 2012, p.
  1752–1759.

\bibitem{tra_deblur2}
T.~Hyun~Kim, K.~Mu~Lee, Generalized video deblurring for dynamic scenes, in:
  The IEEE Conference on Computer Vision and Pattern Recognition (CVPR), 2015.

\bibitem{tra_deblur3}
W.~Ren, J.~Pan, X.~Cao, M.-H. Yang, Video deblurring via semantic segmentation
  and pixel-wise non-linear kernel, in: The IEEE International Conference on
  Computer Vision (ICCV), 2017.

\bibitem{reblur2deblur}
H.~{Chen}, J.~{Gu}, O.~{Gallo}, M.~{Liu}, A.~{Veeraraghavan}, J.~{Kautz},
  Reblur2deblur: Deblurring videos via self-supervised learning, in: 2018 IEEE
  International Conference on Computational Photography (ICCP), 2018, pp. 1--9.
\newblock \href {http://dx.doi.org/10.1109/ICCPHOT.2018.8368468}
  {\path{doi:10.1109/ICCPHOT.2018.8368468}}.

\bibitem{stfan}
S.~Zhou, J.~Zhang, J.~Pan, H.~Xie, W.~Zuo, J.~Ren,
  \href{http://arxiv.org/abs/1904.12257}{Spatio-temporal filter adaptive
  network for video deblurring}, CoRR abs/1904.12257.
\newblock \href {http://arxiv.org/abs/1904.12257} {\path{arXiv:1904.12257}}.
\newline\urlprefix\url{http://arxiv.org/abs/1904.12257}

\bibitem{rdurud}
H.~Sim, M.~Kim, A deep motion deblurring network based on per-pixel adaptive
  kernels with residual down-up and up-down modules, in: The IEEE Conference on
  Computer Vision and Pattern Recognition (CVPR) Workshops, 2019.

\bibitem{cyclegan}
J.-Y. Zhu, T.~Park, P.~Isola, A.~A. Efros, Unpaired image-to-image translation
  using cycle-consistent adversarial networks, in: The IEEE International
  Conference on Computer Vision (ICCV), 2017.

\bibitem{3dconv}
S.~{Ji}, W.~{Xu}, M.~{Yang}, K.~{Yu}, 3d convolutional neural networks for
  human action recognition, IEEE Transactions on Pattern Analysis and Machine
  Intelligence 35~(1) (2013) 221--231.
\newblock \href {http://dx.doi.org/10.1109/TPAMI.2012.59}
  {\path{doi:10.1109/TPAMI.2012.59}}.

\bibitem{darkchannelprior}
J.~Pan, D.~Sun, H.~Pfister, M.-H. Yang, Blind image deblurring using dark
  channel prior, in: The IEEE Conference on Computer Vision and Pattern
  Recognition (CVPR), 2016.

\bibitem{extremechannelprior}
Y.~Yan, W.~Ren, Y.~Guo, R.~Wang, X.~Cao, Image deblurring via extreme channels
  prior, in: The IEEE Conference on Computer Vision and Pattern Recognition
  (CVPR), 2017.

\bibitem{deblurtext}
J.~Pan, Z.~Hu, Z.~Su, M.-H. Yang, Deblurring text images via l0-regularized
  intensity and gradient prior, in: The IEEE Conference on Computer Vision and
  Pattern Recognition (CVPR), 2014.

\bibitem{discriminativeprior}
L.~Li, J.~Pan, W.-S. Lai, C.~Gao, N.~Sang, M.-H. Yang, Learning a
  discriminative prior for blind image deblurring, in: The IEEE Conference on
  Computer Vision and Pattern Recognition (CVPR), 2018.

\bibitem{classprior}
S.~{Anwar}, C.~P. {Huynh}, F.~{Porikli}, Image deblurring with a class-specific
  prior, IEEE Transactions on Pattern Analysis and Machine Intelligence 41~(9)
  (2019) 2112--2130.
\newblock \href {http://dx.doi.org/10.1109/TPAMI.2018.2855177}
  {\path{doi:10.1109/TPAMI.2018.2855177}}.

\bibitem{featureprior}
J.~Peng, Y.~Shao, N.~Sang, C.~Gao,
  \href{http://www.sciencedirect.com/science/article/pii/S0031320320301047}{Joint
  image deblurring and matching with feature-based sparse representation
  prior}, Pattern Recognition 103 (2020) 107300.
\newblock \href
  {http://dx.doi.org/https://doi.org/10.1016/j.patcog.2020.107300}
  {\path{doi:https://doi.org/10.1016/j.patcog.2020.107300}}.
\newline\urlprefix\url{http://www.sciencedirect.com/science/article/pii/S0031320320301047}

\bibitem{flownet2.0}
E.~Ilg, N.~Mayer, T.~Saikia, M.~Keuper, A.~Dosovitskiy, T.~Brox, Flownet 2.0:
  Evolution of optical flow estimation with deep networks, in: The IEEE
  Conference on Computer Vision and Pattern Recognition (CVPR), 2017.

\bibitem{rcan}
Y.~Zhang, K.~Li, K.~Li, L.~Wang, B.~Zhong, Y.~Fu, Image super-resolution using
  very deep residual channel attention networks, in: The European Conference on
  Computer Vision (ECCV), 2018.

\bibitem{perceptualsimilarity}
R.~Zhang, P.~Isola, A.~A. Efros, E.~Shechtman, O.~Wang, The unreasonable
  effectiveness of deep features as a perceptual metric, in: The IEEE
  Conference on Computer Vision and Pattern Recognition (CVPR), 2018.

\bibitem{deblurgan}
O.~Kupyn, V.~Budzan, M.~Mykhailych, D.~Mishkin, J.~Matas, Deblurgan: Blind
  motion deblurring using conditional adversarial networks, in: The IEEE
  Conference on Computer Vision and Pattern Recognition (CVPR), 2018.

\bibitem{firstperson}
G.~Garcia-Hernando, S.~Yuan, S.~Baek, T.-K. Kim, First-person hand action
  benchmark with rgb-d videos and 3d hand pose annotations, in: The IEEE
  Conference on Computer Vision and Pattern Recognition (CVPR), 2018.

\bibitem{frameinterpolation}
S.~Niklaus, L.~Mai, F.~Liu, Video frame interpolation via adaptive separable
  convolution, in: The IEEE International Conference on Computer Vision (ICCV),
  2017.

\end{thebibliography}

\end{document}